\documentclass[letterpaper, 10 pt, conference]{ieeeconf}  

\IEEEoverridecommandlockouts                              

\usepackage{graphics} 
\usepackage{epsfig} 
\usepackage{times} 
\usepackage{amsmath} 
\usepackage{amssymb}  
\usepackage{cite}
\usepackage{subfigure}
\usepackage{xcolor}
\usepackage{caption, bm}

\usepackage{hyperref}

\title{\LARGE \bf
Incremental Learning for Robot Shared Autonomy}

\author{Yiran Tao$^{1}$, Guixiu Qiao$^{2}$, Dan Ding$^{3}$, Zackory Erickson$^{1}$
\thanks{$^{1}$Robotics Institute, Carnegie Mellon University, Pittsburgh, PA, USA}
\thanks{$^{2}$National Institute of Standards and Technology, Gaithersburg, MD, USA}%
\thanks{$^{3}$Department of Rehabilitation Science and Technology, School of Health and Rehabilitation Sciences, University of Pittsburgh, Pittsburgh, PA, USA}
\thanks{This material is based upon work supported by NIST under Grant No. 70NANB23H216.}
}

\begin{document}

\maketitle
\thispagestyle{empty}
\pagestyle{empty}

\begin{abstract}

Shared autonomy holds promise for improving the usability and accessibility of assistive robotic arms, but current methods often rely on costly expert demonstrations and remain static after pretraining, limiting their ability to handle real-world variations. Even with extensive training data, unforeseen challenges—especially those that fundamentally alter task dynamics, such as unexpected obstacles or spatial constraints—can cause assistive policies to break down, leading to ineffective or unreliable assistance. To address this, we propose ILSA, an Incrementally Learned Shared Autonomy framework that continuously refines its assistive policy through user interactions, adapting to real-world challenges beyond the scope of pre-collected data. At the core of ILSA is a structured fine-tuning mechanism that enables continual improvement with \textit{each} interaction by effectively integrating limited new interaction data while helping to preserve prior knowledge, aiming for a balance between adaptation and generalization. A user study with 20 participants demonstrates ILSA’s effectiveness, showing faster task completion and improved user experience compared to static alternatives. Code and videos are available at \url{https://ilsa-robo.github.io/}.

\end{abstract}


\section{INTRODUCTION}
Assistive devices such as wheelchair-mounted robotic arms have the potential to benefit millions of individuals with disabilities and enhance their independence~\cite{wheelchair1, wheelchair2}. 
Shared autonomy, where human operators and autonomous systems share decision-making, has emerged as a promising way to make these devices more intuitive and effective~\cite{safe1, safe2}.

\begin{figure}[tbp]
\centering

\subfigure[Cereal pouring task, trial 1]{
\begin{minipage}[t]{0.42\columnwidth}
\centering
\includegraphics[width=\columnwidth]{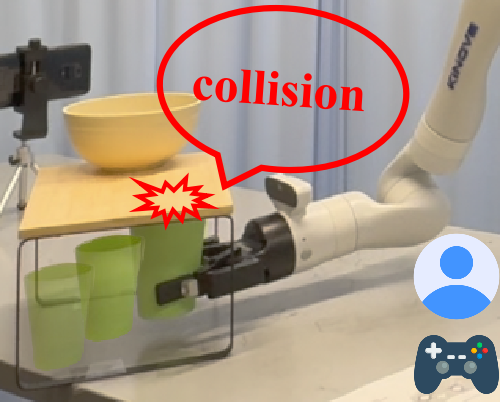}
\end{minipage}%
}%
\subfigure[Cereal pouring task, trial 4]{
\begin{minipage}[t]{0.42\columnwidth}
\centering
\includegraphics[width=\columnwidth]{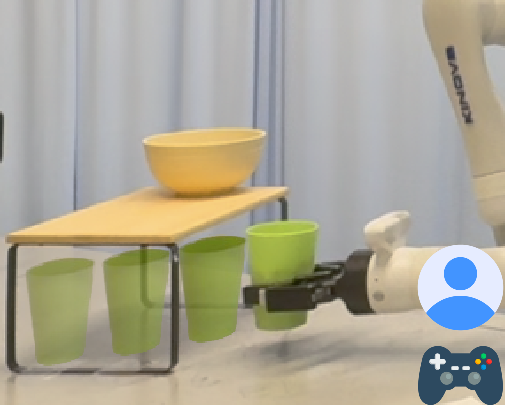}
\end{minipage}%
}%
\quad                 
\subfigure[Pill bottle storage task, trial 1]{
\begin{minipage}[t]{0.42\columnwidth}
\centering
\includegraphics[width=\columnwidth]{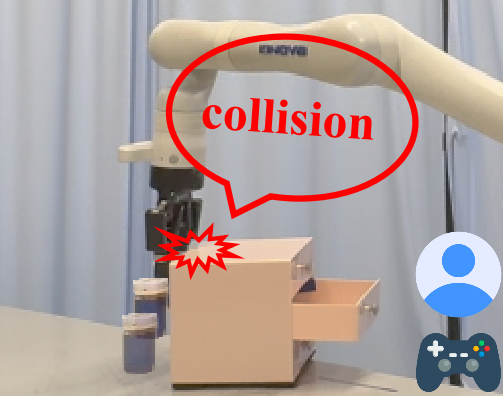}
\end{minipage}
}%
\subfigure[Pill bottle storage task, trial 4]{
\begin{minipage}[t]{0.42\columnwidth}
\centering
\includegraphics[width=\columnwidth]{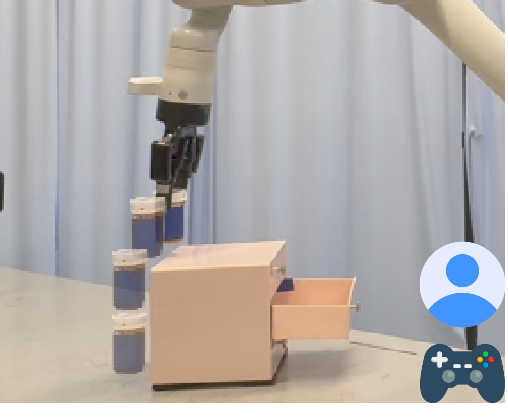}
\end{minipage}
}%
\centering
\caption{We introduce ILSA, an Incrementally Learned Shared Autonomy system that continuously refines its assistive policy through user interaction, enabling adaptation to real-world challenges beyond the scope of pre-collected training data. \textbf{Left}: In early trials, ILSA struggles with collision avoidance due to obstacles or spatial constraints not accounted for in training data. \textbf{Right}: With only a few interactions, by the fourth trials, ILSA has learned effective collision avoidance from past user interactions, assisting the user in completing tasks more smoothly.}
\label{fig1}
\vspace{-1em}
\end{figure}

Recent shared autonomy systems have begun employing imitation learning, which learns an assistive control policy from human demonstrations to enhance generalization across task environments~\cite{latent, latentuser, lila, no, noise}. However, collecting large-scale demonstrations is costly and difficult to scale~\cite{survey1, survey2, survey3}. Moreover, even with extensive training data, it is impractical to cover the full range of real-world scenarios. As a result, assistive policies trained solely on pre-collected data may struggle when encountering previously unseen situations during deployment. This limitation becomes particularly critical when the training data fails to anticipate specific real-world challenges that fundamentally alter the task dynamics, such as the presence of obstacles or spatial constraints. For example, in a cereal pouring task where the system needs to move a cereal container next to a bowl to pour, as shown in the upper two sub-figures of Fig.~\ref{fig1}, even a well-designed training set may not account for all possible real-world variations, such as the presence of obstacles between the two objects. If such obstacles exist in deployment, a policy trained without exposure to these cases may struggle to adapt, leading to suboptimal or even infeasible behavior.

To address the inability of static policies to adapt to unexpected real-world challenges that alter task execution, we introduce an Incrementally Learned Shared Autonomy (ILSA) framework (Fig.~\ref{fig:model}) that continuously fine-tunes its assistive policy as the user interacts with the system. A key technical contribution of ILSA is its structured fine-tuning mechanism, which updates the assistive policy after \textit{each} user interaction, ensuring enhanced assistance in \textit{every} subsequent use. This frequent adaptation is particularly challenging due to limited deployment data, which can affect model stability and generalizability. To address this, our fine-tuning mechanism consists of three key components: (1) \textbf{Corrected Trajectory Supervision}, which prevents the model from reinforcing suboptimal actions by generating corrected trajectories from user interactions and using them for fine-tuning; (2) \textbf{Layered Supervision}, which integrates new interaction data with pretraining knowledge to balance adaptation and retention, without requiring manual data reweighting; and (3) \textbf{Partial Model Update}, which selectively fine-tunes critical components to enhance adaptation while maintaining generalization and robustness. These designs collectively enable continual adaptation while helping to preserve previously acquired knowledge, thus supporting ILSA’s generalization ability gained from pretraining. Our fine-tuning approach is validated through a quantitative ablation study.

While our fine-tuning mechanism enables ILSA to continuously adapt to user interactions, the initial pretraining phase remains essential for ensuring effective first-time use. Instead of relying on costly demonstrations that still cannot capture all real-world challenges, we initialize ILSA with a lightweight set of simulated kinematic trajectories, generated through simple rule-based procedures. This allows for efficient task initialization while minimizing training costs.

To formally evaluate ILSA, we conducted a user study with 20 participants using a Kinova robotic arm to perform two complex, long-horizon tasks. These tasks were specifically designed to introduce challenges that were not accounted for in the training data, such as the presence of obstacles or spatial constraints, highlighting scenarios where pre-collected demonstrations may fail to generalize.
Crucially, we note that these challenges are not a consequence of using simulated data but rather reflect the broader issue that even expert demonstrations cannot account for every possible variation.
Our user study results show that: (1) ILSA enables faster task completion and more intuitive manipulator control compared to pure teleoperation; (2) ILSA effectively adapts to real-world challenges beyond the scope of training data, continuously improving task performance over time, in contrast to a static shared autonomy method.


Our contributions are summarized as follows:

\begin{itemize}
  \item We propose ILSA, an Incrementally Learned Shared Autonomy framework that continuously improves its assistive policy through user interaction, enabling adaptation to real-world challenges beyond the scope of pre-collected data.
  \item We design a structured fine-tuning mechanism that enables ILSA to improve its assistive policy after \textit{each} user interaction, helping to achieve continual improvement while balancing adaptation and generalization ability. We validate its effectiveness through a quantitative ablation study.
  \item We conducted a human study, demonstrating that ILSA rapidly enhances task performance over time, significantly outperforming teleoperation and non-incremental shared autonomy in real-world tasks with previously unseen challenges. 
\end{itemize}

\section{RELATED WORK}
\subsection{Shared Autonomy with Continual Adaptation}

In recent years, a limited number of shared autonomy works explored frameworks that refine assistive policies through repeated user interactions. Two notable works in this direction are Confidence-Aware Shared Autonomy (CASA)~\cite{casa} and Shared Autonomy across Repeated Interactions (SARI)~\cite{sari}. Both methods aim to improve assistive policies over time by expanding the robot’s skill set through learning new tasks or goals from user demonstrations. Specifically, CASA estimates confidence in a user’s intent to determine whether a task is known or novel. If a task is unrecognized, it relinquishes control entirely, allowing the user to demonstrate the task. CASA then applies inverse reinforcement learning (IRL) and incorporate the new task into its skill set. Similarly, SARI automates familiar tasks based on prior interactions while deferring control to the user for novel ones. It periodically retrains its policy on collected trajectories, enabling continual refinement over time.


While these approaches enhance assistance across repeated interactions, they focus on expanding the robot’s skill set by learning new tasks rather than adapting within an existing one. Once a task is learned, they assume it remains unchanged. In contrast, our work addresses a different but equally critical challenge: even within a single task, real-world variations—especially those that fundamentally alter task dynamics, such as unforeseen obstacles or spatial constraints—can significantly impact task execution, requiring continuous adaptation.

Furthermore, CASA and SARI require multiple user demonstrations before updating and retrain periodically on aggregated interaction data, which can lead to long periods of suboptimal performance before policy updates, limiting the user experience. In contrast, we aim to fine-tune the assistive policy after \textit{each} interaction, enabling continuous adaptation in real-time. To achieve this, we focus on effectively integrating new interaction data while preserving pretraining knowledge, striking a balance between adaptation and retention. While prior shared autonomy frameworks have not explicitly explored this direction, we introduce a structured fine-tuning mechanism specifically designed for this challenge. Through our key designs, ILSA ensures that each update contributes to improved assistance, effectively enabling adaptation to real-world challenges beyond the scope of pre-collected data with minimal interaction overhead.

\begin{figure*}[t]
\centering
\includegraphics[width=2\columnwidth]{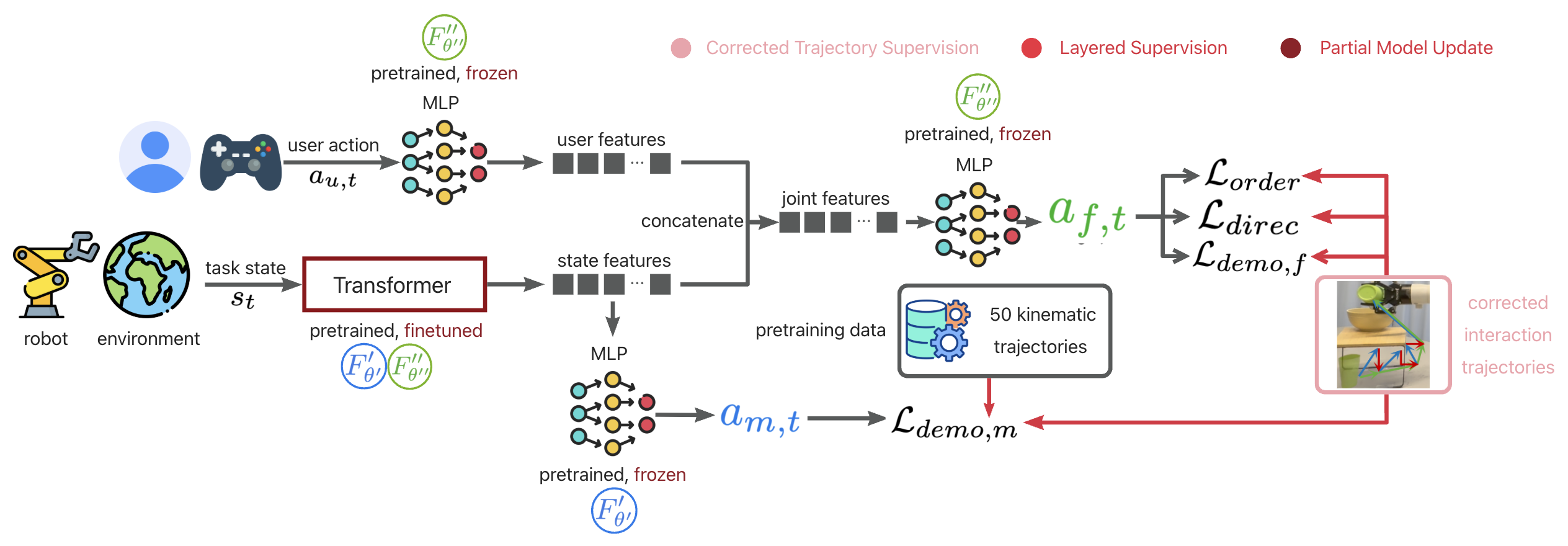} 
\caption{Our Incrementally Learned Shared Autonomy (ILSA) framework. The base action generation model $\mathcal{F}_{\theta}(s_{t}, a_{u,t})$ consists of an intermediate action predictor $\mathcal{F'}_{\theta'}(s{t})$ (blue) and a final action predictor $\mathcal{F''}_{\theta''}(s_{t}, a_{u,t})$ (green). At the core of ILSA is its structured fine-tuning mechanism (highlighted in red), which incorporates three key components: Corrected Trajectory Supervision, Layered Supervision, and Partial Model Update. These designs enable ILSA to continuously improve with \textit{every} user interaction while preserving generalization ability, effectively adapting to real-world challenges beyond pre-collected training data.}
\label{fig:model}
\end{figure*}

\subsection{Human-in-the-loop Machine Learning}
\label{sec:human}
Another related area is human-in-the-loop machine learning~\cite{hil00, hil01}, where autonomous agents refine their policies through human interactions. A notable example is DAgger (Dataset Aggregation)\cite{dagger} and its variants, such as EnsembleDAgger\cite{ensemble} and HG-DAgger~\cite{hg-dagger}, which address covariate shift in imitation learning by iteratively collecting expert interventions on novel states encountered during deployment, improving policy robustness.

While effective for fully autonomous systems, these methods rely on expert interventions, whereas our work focuses on shared autonomy, where real-time assistance is needed for non-expert users. Furthermore, while human-in-the-loop learning has been shown to improve policy performance, existing approaches primarily evaluate adaptation to common task variations—such as differences in object positions or trajectories—rather than situations that fundamentally alter task execution, like unforeseen obstacles or spatial constraints. On the contrary, ILSA is motivated by the need for adaptation in these challenging scenarios and demonstrates its ability to handle them in our experiments.

Additionally, human-in-the-loop frameworks typically require multiple policy deployments and substantial expert feedback before updating the model, leading to extended periods of suboptimal performance. In contrast, ILSA specifically addresses the challenge of per-interaction adaptation, introducing a structured fine-tuning mechanism that seamlessly integrates new interaction data while preserving pretraining knowledge. This design enables stable and generalizable adaptation even with limited data, ensuring continuous improvement after each user interaction.


\section{Methodology}
\subsection{Problem Statement}
We formulate our shared autonomy task as a sequential decision process defined by tuple $(\mathcal{S}, \mathcal{A}_{u}, \mathcal{A}_{r}, \mathcal{T})$, where $s_{t} \in \mathcal{S}$ denotes the task state at time step $t$, $a_{u,t} \in \mathcal{A}_{u}$ denotes the user action, $a_{r,t} \in \mathcal{A}_{r} $ denotes the robot action (we use the 6-DoF delta end-effector position, including Cartesian coordinates and Euler angles). $\mathcal{T}$: $\mathcal{S} \times \mathcal{A}_{r} \rightarrow \mathcal{S}$ is an unobserved transition function.

Our goal is to identify a function $\mathcal{F}_{\theta}(s_{t}, a_{u,t})$: $\mathcal{S} \times \mathcal{A}_{u} \rightarrow \mathcal{A}_{r}$ that maps the task state and user action to a robot action that is both task-optimal and aligned with the user's intent. In our work, we model $\mathcal{F}_{\theta}$ using a robot action generation model (Fig.~\ref{fig:model}). Unlike traditional methods that solely rely on static pre-collected demonstrations to train this model, we adopt a two-stage learning process: 

1) \textit{Pretraining} with lightweight, simulated kinematic trajectories to provide an initial assistive policy. 

2) \textit{Incremental adaptation }through real-world user interactions to continually refine the policy and address unforeseen deployment challenges.

In the following sections, we first describe our simulated trajectory generation process (Section~\ref{sec:simulated}), then introduce our robot action generation model (Section~\ref{sec:model}), which serves as the basis of ILSA’s assistive policy. Finally, we present our structured fine-tuning mechanism (Section~\ref{sec:incre}), the key technical contribution that addresses challenges such as learning from sparse deployment data and maintaining generalization ability, enabling ILSA to continuously adapt to user interactions.

\subsection{Simulated Kinematic Trajectories Generation}
\label{sec:simulated}
Instead of relying on costly expert demonstrations, which are still unlikely to cover all real-world variations, we use a simple, rule-based approach to generate simulated kinematic trajectories to pretrain ILSA. 

To generate these trajectories, we decompose multi-step tasks into distinct subtasks based on predefined task objectives. Each subtask corresponds to a critical transition in task execution, such as object grasping, transport, or tool manipulation. For instance, in a cereal-pouring task consisting of three subtasks—(1) grasping the cereal container, (2) moving it next to the bowl, and (3) pouring—the container and bowl positions are randomly sampled within the workspace, and the corresponding end-effector poses (positions and orientations) necessary to complete each subtask are computed.
The robot's motion is then computed via linear interpolation between initial and target end-effector poses, forming state-action tuples $(s_t, a_{r,t})$. Here, $s_t$ includes the full task state, encompassing the robot’s end-effector pose as well as the poses of all relevant objects in the scene.


To further simulate user interaction, we generate synthetic user actions $a_{u,t}$ to mimic user inputs. The user action is defined as a 3D vector representing intended end-effector movements in the $x$, $y$, and $z$ directions. Each dimension corresponds to a normalized velocity command: if the target position in dimension $i$ is greater than the current position, we sample $a_{u,t,i}$ from $[0,1]$; otherwise, we sample from $[-1,0]$. In real-world deployment, these inputs are provided by a user via a dual-joystick Xbox controller, where one joystick operates two translational axes (e.g., $x$ and $y$), and the other controls the third axis ($z$). Note that for usability, in most cases, users control only translation (3D) while the robot autonomously generates the end-effector orientation. When direct rotational control is necessary, the user's input bypasses the policy and is executed directly. This design reduces user burden and was validated in pilot studies.

We generate 50 trajectories per task to ensure sufficient diversity for pretraining. Each trajectory consists of a sequence of $(s_t, a_{u,t}, a_{r,t})$ triplets. This lightweight, automated data generation strategy ensures an effective initialization. The simplicity of the trajectory generation process also facilitates easy transferability across different manipulation tasks without requiring complex environment-specific modeling.

\subsection{Robot Action Generation Model}
\label{sec:model}

Our robot action generation model $\mathcal{F}_{\theta}(s_{t}, a_{u,t})$, shown in Fig.~\ref{fig:model}, produces two types of actions at each time step: an intermediate action $a_{m,t}=\mathcal{F'}_{\theta'}(s_{t})$ without user guidance and a final action $a_{f,t}=\mathcal{F''}_{\theta''}(s_{t}, a_{u,t})$ with user guidance. Here, $\mathcal{F'}_{\theta'}$ and $\mathcal{F''}_{\theta''}$ represent different subsets of the overall model $\mathcal{F}_{\theta}$, as illustrated in Fig.~\ref{fig:model}.While only the action with user guidance is executed, having an intermediate output is crucial during our incremental learning process (Section~\ref{sec:incre}). 

To pretrain $\mathcal{F}_{\theta}$, we use a set of simulated kinematic trajectories (Section~\ref{sec:simulated}) and optimize a loss function designed to balance task completion and user intent alignment. This function includes terms ($\mathcal{L}_{demo, m}$, $\mathcal{L}_{demo, f}$) that align both intermediate and final actions with training data to ensure task completion, along with additional constraints ($\mathcal{L}_{direc}$, $\mathcal{L}_{order}$) to maintain consistency with user inputs. Since this model serves as the basis for ILSA’s assistive policy rather than its core technical contribution, which focuses on incremental adaptation to real-world challenges, we defer the full implementation details—including network architecture and training configuration—to Appendix~\ref{appendix:implementation}.

\subsection{Structured Incremental Learning Mechanism}
\label{sec:incre}
To enable ILSA to adapt to unforeseen real-world challenges beyond pre-collected data, we introduce a structured incremental learning mechanism.

This process begins with how ILSA interacts with the user in real time: During deployment, at each time step, the user provides an action $a_{u,t}$ through the control interface, which is input into $\mathcal{F}_\theta$ to generate the final robot action $a_{f,t}$. Since real-world deployment may introduce unforeseen challenges beyond the scope of pretraining, the pretrained model may generate suboptimal actions when encountering these novel situations. To address this, we monitor the alignment between user inputs and model predictions by computing the cosine similarity between $a_{u,t}$ and $a_{f,t}$. If the similarity exceeds a threshold (empirically set to 0.5), $a_{f,t}$ is executed. Otherwise, we alert the user by briefly pausing the robot’s motion. If the user persists with the same input, the robot will execute $a_{u,t}$ instead of $a_{f,t}$. This process generates new interaction data, which we use to incrementally update $\mathcal{F}_{\theta}$.

While a straightforward approach would be to directly use these interaction trajectories for fine-tuning, this presents several issues: (1) suboptimal trajectories introduced by initial model limitations may be reinforced rather than corrected, (2) adapting to new scenarios risks forgetting of previously learned behaviors, and (3) learning from limited deployment data can lead to instability or reduce generalization ability. To address these challenges, we propose three key designs: Corrected Trajectory Supervision, Layered Supervision, and Partial Model Update, which together enable robust and efficient incremental learning.

\subsubsection{Corrected Trajectory Supervision}

Our experiments show that when users overwrite $a_{f,t}$ with $a_{u,t}$, it is typically to correct suboptimal actions previously made by the system, which is usually caused by unforeseen task variations. For example, as illustrated in Fig.~\ref{fig1}(a), the shelf above the cup is an unexpected obstacle that the model was not exposed to during training, leading the robot arm to collide with it while attempting to retrieve the cup. To recover, the user intervenes by commanding the robot to move downward, an action that deviates from the model’s expected behavior and requires overwriting $a_{f,t}$ with $a_{u,t}$. This correction is only necessary because the initial model, unaware of the obstacle, attempted an infeasible trajectory that goes up.

As illustrated in this example, raw trajectories contain both suboptimal policy actions and user corrections that wouldn't be necessary without the suboptimal actions. Directly using these trajectories for training risks misleading the model into treating these corrective actions as optimal behavior.

\begin{figure}[htbp]
\centering
\includegraphics[width=0.6\columnwidth]{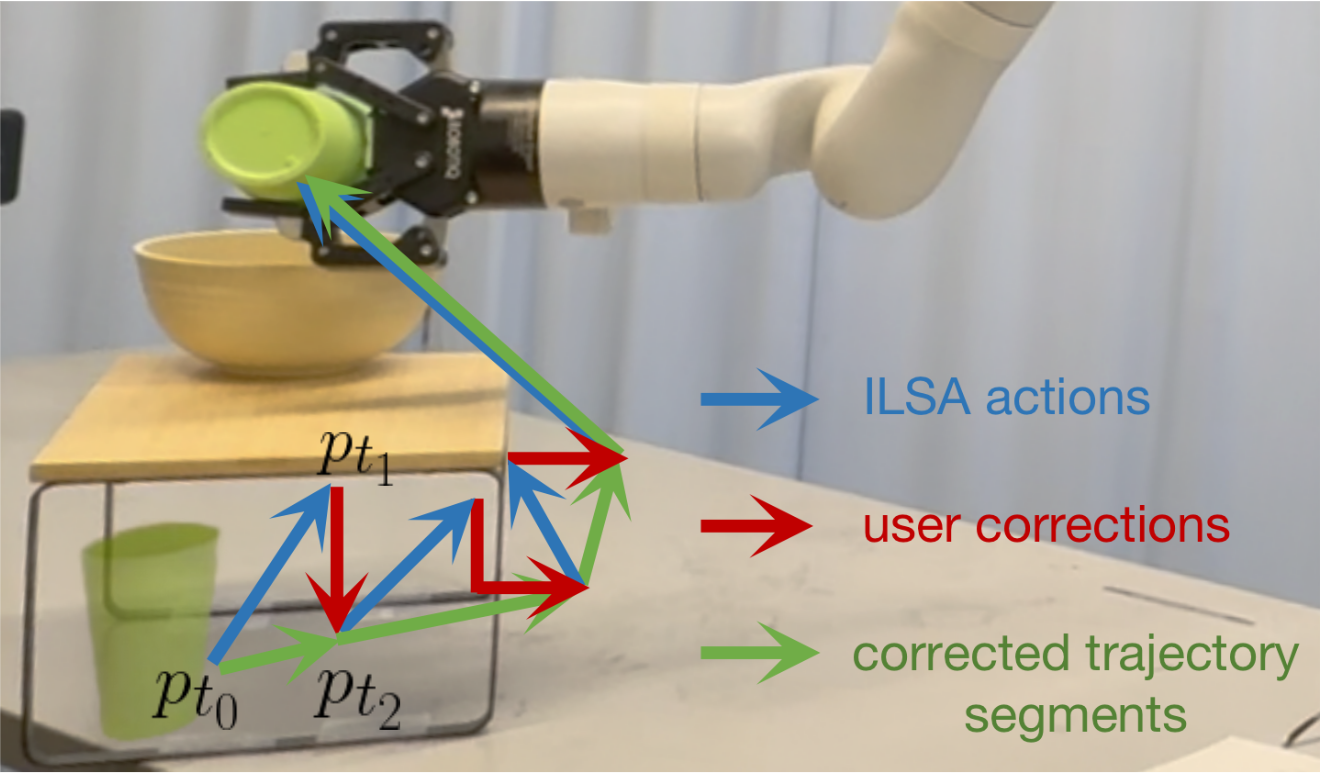} 
\caption{A corrected trajectory (green segments) constructed from ILSA actions (blue segments) and user corrections (red segments) for the first trial of a task.}
\label{fig:correction}
\end{figure}

To address this, we generate a ``corrected trajectory'' for each raw trajectory observed during human use. As shown in Fig.~\ref{fig:correction}, when the user overwrites $a_{f,t}$ (red segments in Fig.~\ref{fig:correction}), let $t_1$ be the start of the overwrite and $t_2$ the time when the robot resumes executing $a_{f,t}$. Before $t_1$, if the robot executes $a_{f,t}$ for a period (blue segments), let $t_0$ be the start of that period. Let $p_{t_0}$, $p_{t_1}$, and $p_{t_2}$ represent the robot's positions at these times. 

Our assumption is that by correcting the robot position from $p_{t_1}$ to $p_{t_2}$, the user essentially believes the ideal trajectory goes directly from $p_{t_0}$ to $p_{t_2}$. Therefore, we simply take the linear trajectory between $p_{t_0}$ and $p_{t_2}$ (green segments) as the corrected trajectory segment. We then perform linear interpolation to generate new $(s_{t}, a_{r,t})$ pairs and apply the same method from Section~\ref{sec:simulated} to generate $a_{u,t}$, forming new $(s_{t}, a_{u,t}, a_{r,t})$ triplets for training.

\subsubsection{Layered Supervision}
\label{sec:layered}

To ensure a smooth user experience, ILSA updates its policy after \textit{each} interaction. However, frequent updates mean that each training step relies on only a small set of newly collected data, making the system susceptible to overfitting and catastrophic forgetting of previously acquired knowledge~\cite{cata1, small3}. 

A naïve solution is to mix newly collected trajectories with the pretraining trajectories. However, this approach is problematic due to the severe data imbalance—especially in early updates when only a few new trajectories exist. A common alternative used in human-in-the-loop machine learning methods is to assign different weights to different types of data~\cite{feifei, job, job2}, but determining optimal weights is nontrivial and often task-specific.

Instead, we propose a \textbf{layered supervision} strategy that naturally integrates both new interaction data and simulated pretraining data without explicit reweighting:
\begin{enumerate}
    \item \textbf{Final Output Supervision}: The new user-generated trajectories are used exclusively to supervise the final output action $a_{f,t}$, ensuring adaptation to new data.
    \item \textbf{Intermediate Output Supervision}: Both simulated kinematic trajectories and new user-generated trajectories supervise the intermediate action $a_{m,t}$, maintaining the model’s generalization ability while still allowing adaptation.
\end{enumerate}

This approach, illustrated by the red arrows in Fig.~\ref{fig:model}, helps ILSA remain adaptive while mitigating the risk of catastrophic forgetting and preserving knowledge from pretraining. Compared to explicit weighting schemes, layered supervision provides a more principled way to balance adaptation and generalization. 

\subsubsection{Partial Model Update}
To further enhance robustness, we update only a subset of $\mathcal{F}_{\theta}$ during fine-tuning. Specifically, we focus on the \textbf{transformer component}, which processes the task state $s_{t}$, while freezing other network modules. This design is motivated by the need for ILSA to efficiently adapt to previously unseen real-world variations—such as new obstacles or spatial constraints—without discarding valuable prior knowledge. Since the transformer extracts high-level task representations from $s_{t}$, fine-tuning it alone allows the policy to evolve with each user interaction while preserving generalization ability. Our ablation study (Section~\ref{sec:ablation}) demonstrates that this selective update strategy significantly outperforms full-model updates or modifying other components.

In each incremental update, we fine-tune the model for 10 epochs on the new and pretraining data, with this value empirically set for stable adaptation.

\section{Experiments}
\subsection{Tasks and Experiment Settings}
All experiments are conducted with a 6-DOF Kinova Gen3 robotic arm.

As described in Section~\ref{sec:simulated}, our pretraining phase relies on simulated kinematic trajectories that do not account for real-world complexities such as obstacles or spatial constraints. However, in real-world applications, previously unseen factors can fundamentally alter task execution. To evaluate ILSA’s ability to adapt in such scenarios, we design our experimental tasks to introduce obstacles and spatial constraints that significantly impact motion feasibility.

Specifically, we design two complex, multi-stage tasks to test the effectiveness of our model:

\subsubsection{Cereal Pouring} 
Pick up a cup of cereal placed under a shelf and pour the cereal into a bowl on top of the shelf. 

This task introduces an obstacle (the shelf) that was not accounted for in pretraining.

\subsubsection{Pill Bottle Storage} 
Open a drawer, pick up a pill bottle behind the drawer, and place it back inside.  

In this task, the spatial constraints between the drawer and the pill bottle were not accounted for in pretraining, requiring ILSA to adapt its motion strategy to avoid collisions.

For each task, a complete experiment consists of four interactions, where the user completes the task four times with different object positions each time. We refer to these interactions as ``trial 1'', ``trial 2'', ``trial 3'', and ``trial 4''. ILSA updates $\mathcal{F}_{\theta}$ after each trial.

\begin{figure}[htbp]
\centering
\includegraphics[width=\columnwidth]{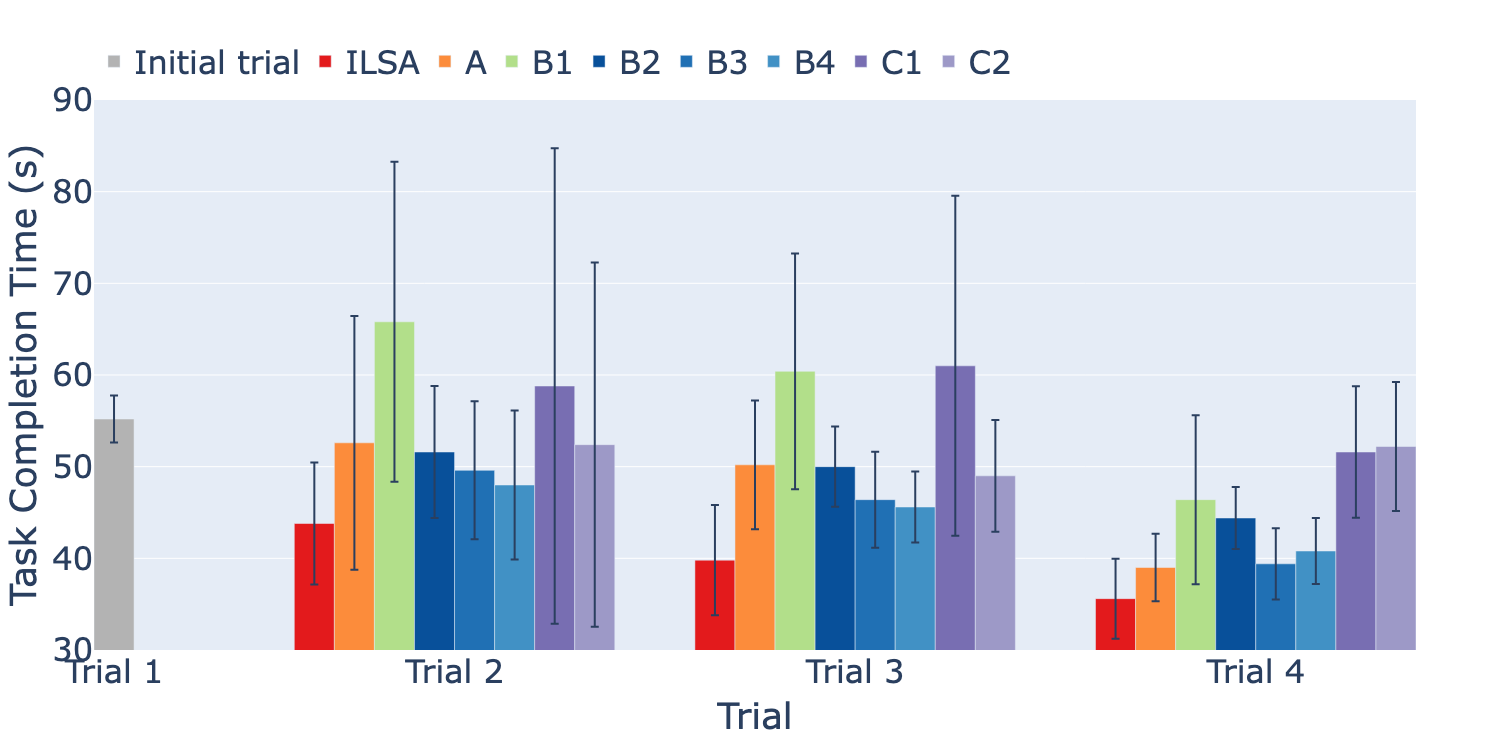} 
\caption{Average task completion times from our ablation study on the cereal pouring task. Error bars show standard deviations.}
\label{fig:ablation}
\vspace{-1em}
\end{figure}

\subsection{Ablation Study}
\label{sec:ablation}
Before conducting the user study, we performed an ablation study to systematically evaluate ILSA’s structured fine-tuning mechanism. Specifically, we analyzed the impact of its three key components—Corrected Trajectory Supervision, Layered Supervision, and Partial Model Update. We evaluate the impact of these components through the following variations:

\begin{itemize}
  \item \textbf{Corrected Trajectory Supervision:} \textbf{A.} Fine-tune on original uncorrected trajectories.
  \item \textbf{Layered Supervision} \textbf{B1.} Use new data only to supervise both $a_{f,t}$ and $a_{m,t}$; \textbf{B2.} Use both new data and pretraining data to supervise both $a_{f,t}$ and $a_{m,t}$; \textbf{B3.} Use new data and randomly selected 10 pretraining trajectories to supervise both $a_{f,t}$ and $a_{m,t}$; \textbf{B4.} Use both new data and pretraining data to supervise both $a_{f,t}$ and $a_{m,t}$, reweighted with $w_{new} = 0.5\cdot \frac{|D_{new}|+|D_{kinematic}|}{|D_{new}|}$ and $w_{kinematic} = 0.5\cdot \frac{|D_{new}|+|D_{kinematic}|}{|D_{kinematic}|}$, where $|D_{new}|$ and $|D_{kinematic}|$ represent the number of data points in the new and kinematic trajectories, respectively (This design is inspired by human-in-the-loop machine learning works \cite{feifei, job}).
  \item \textbf{Partial Model Update} \textbf{C1.} Update the entire architecture; \textbf{C2.} Freeze the transformer and update other components.
\end{itemize}

For the ablation study, one researcher conducted five experiments for each model variant on the cereal pouring task. The tested designs were randomized and interleaved across experiments. Object positions were randomized for each trial, resulting in a total of 20 unique task layouts across the five experiments. All model variants were evaluated on the same sequence of object positions within each experiment. This ensures that performance differences arise from model adaptation rather than variations in task difficulty. For all methods, after each trial, we fine-tuned for a fixed number of epochs, with the value empirically set for each variant to ensure stable adaptation.

Across both the ablation and user studies (Section~\ref{sec:human}), we primarily \footnote{Additional experimental results are available on our project website.} evaluate task completion time as a direct measure of user efficiency and system adaptability.

The task completion times of different models are shown in Fig.~\ref{fig:ablation}. As shown in the figure, our proposed method, ILSA, consistently decreases task completion times across trials. With just a few user interactions, ILSA lowers the average task completion time from 55.2 seconds in trial 1 to 35.6 seconds in trial 4, which is a 35.5\% reduction, suggesting effective adaptation to unforeseen obstacles that were not accounted for during pretraining. From trial 2 to trial 4, ILSA has the lowest average task completion time and one of the smallest standard deviations, demonstrating both effectiveness and repeatability.

In contrast, the method using uncorrected trajectories (A1) shows a smaller reduction in task completion time in trials 2 and 3, with larger standard deviations. This is mainly because user corrections for previous suboptimal actions in the uncorrected trajectories are outliers that can cause overfitting. This can lead to errors in later trials, resulting in higher average task completion times and standard deviations.

Similarly, updating the model only on new data (B1), updating the entire architecture (C1), or freezing the transformer and updating other components (C2) also leads to worse incremental learning and task performance. These ablated methods performed poorly on task layouts that are highly dissimilar to previously seen scenarios.

The three methods that use both new data and simulated kinematic trajectories to supervise both $a_{f,t}$ and $a_{m,t}$ (B2, B3, B4) show a relatively steady decrease in average task completion times. However, compared to ILSA, they improve performance at a slower rate, indicating that they are not utilizing new data as efficiently as ILSA.

\begin{figure}[htbp]
\centering

\subfigure[Cereal pouring task]{
\begin{minipage}[t]{\columnwidth}
\centering
\includegraphics[width=\columnwidth]{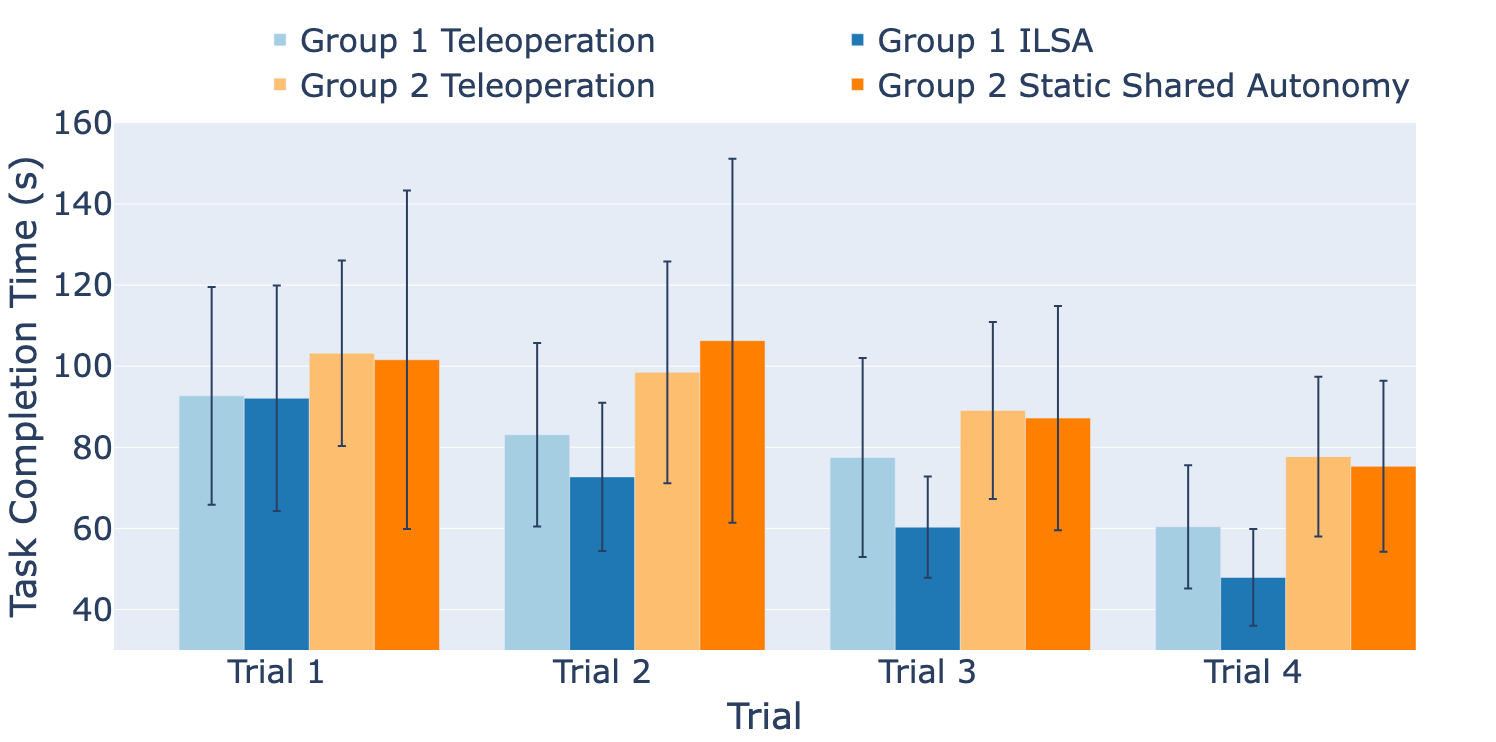}
\end{minipage}%
}%
\quad                
\subfigure[Pill bottle storage task]{
\begin{minipage}[t]{\columnwidth}
\centering
\includegraphics[width=\columnwidth]{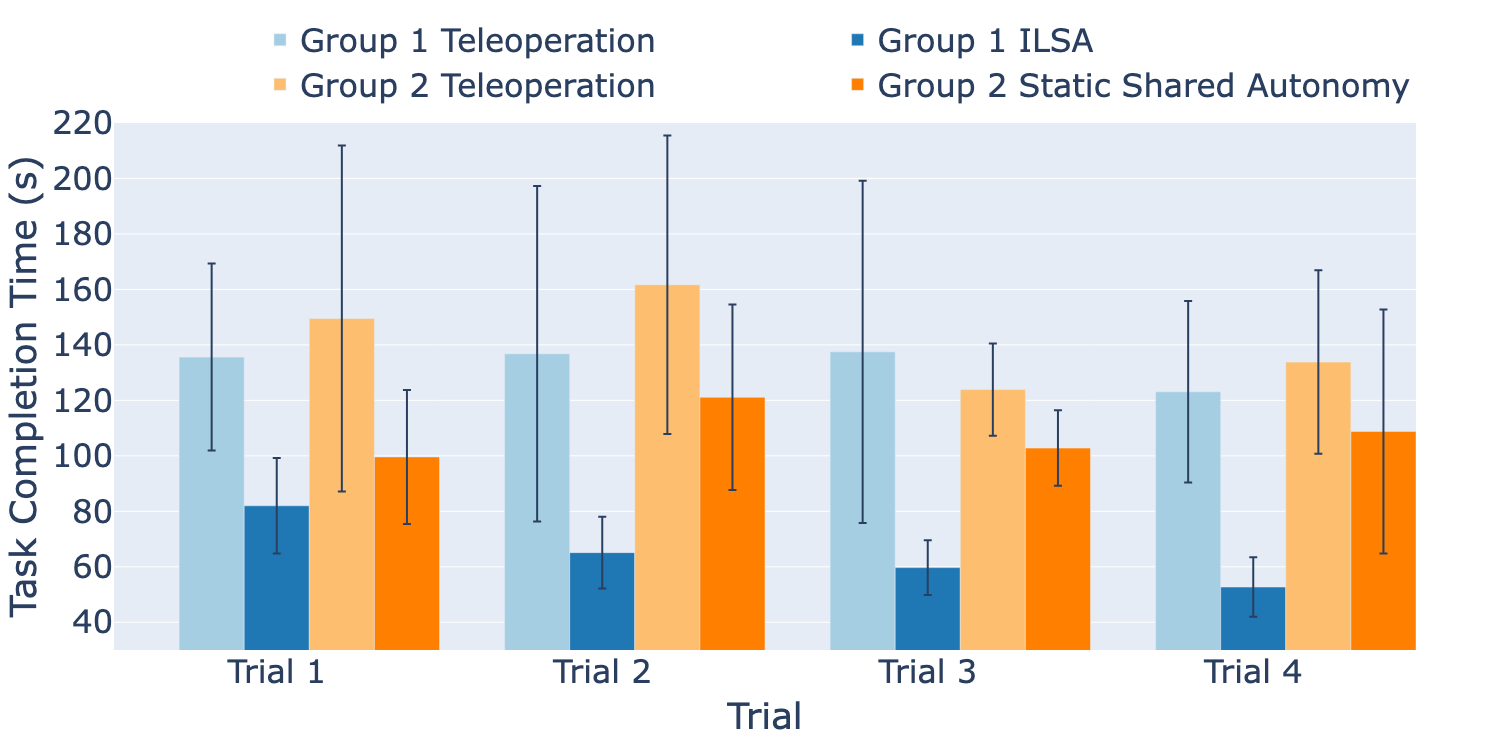}
\end{minipage}
}%

\centering
\caption{Task completion times averaged over all participants in each group. Error bars show standard deviations.}
\label{fig:time}
\vspace{-1em}
\end{figure}

\subsection{User Study}
To formally evaluate ILSA, We conducted a user study subject to a university-approved IRB protocol with 20 participants, none of whom reported having prior experience teleoperating robots. We test the following two hypotheses:
\begin{itemize} 
  \item \textbf{H1:} ILSA supports faster robot manipulation task completion times and easier manipulator control as compared to pure teleoperation.
  \item \textbf{H2:} ILSA effectively adapts to real-world challenges beyond the scope of training data, continuously improving task performance over time, in contrast to a static shared autonomy method.
\end{itemize}

Participants were divided into two groups. Group 1, consisting of 10 participants, used ILSA and pure teleoperation, while Group 2, consisting of 10 participants, used teleoperation and a static shared autonomy policy, which was pretrained on the initial simulated kinematic trajectories and kept static across the four task trials. 

Participants completed the task trials twice with different methods in a counterbalanced order. While the within-subject comparison in Group 1 is sufficient to provide evidence towards H1, Group 2 is included to ensure that the reduction in task completion time with ILSA is not simply due to the user's increasing familiarity with the shared autonomy system, which provides support for H2.

\subsubsection{\textbf{Task Completion Times}}

The task completion times of both groups using the varying methods are shown in Fig~\ref{fig:time}. For both tasks across all four trials, ILSA (dark blue bars) allowed for faster task completion in comparison to teleoperation (light blue bars). On average across all four trials, the cereal pouring task with ILSA took 10.2 seconds (12.97\%) less than with teleoperation, while the pill bottle storage task with ILSA took 68.375 seconds (51.31\%) less than with teleoperation. 

A Wilcoxon signed-rank test on the combined task completion times from all four trials shows a statistically significant difference between ISLA and teleoperation ($p < 0.001$ for both the cereal pouring and pill bottle storage tasks), providing evidence towards H1.

In addition, as shown in Fig~\ref{fig:time}, ILSA steadily reduced task completion times for both tasks. After only 3 trials of cereal pouring, ILSA decreased the average completion time from 92.1 seconds to 47.9 seconds, a 48.0\% reduction. For the pill bottle storage task, the average time dropped from 82.0 seconds to 52.7 seconds, a 35.7\% reduction, across all participants in Group 1.

To ensure that the reduction in task completion times with ILSA is not simply due to users' increasing familiarity with the shared autonomy system, we compared results from group 1 with group 2. A Mann-Whitney U test on trial 4 shows no statistically significant difference between teleoperation in group 1 and group 2 ($p > 0.05$ for both tasks). However, there is a statistically significant difference between ILSA and the static shared autonomy policy ($p < 0.01$ for the cereal pouring task and $p < 0.001$ for the pill bottle storage task), supporting H2.

\subsubsection{\textbf{User-reported Qualitative Metrics}}

In addition to task completion time, we collected participants' subjective evaluations using a custom-designed 7-point Likert-scale questionnaire (1 = strongly disagree; 4 = neutral; 7 = strongly agree) to assess their perception of robot control. The evaluated metrics included ``feel in control", ``low mental effort", ``intuitiveness", ``ease of task completion", ``smooth motions", ``precise motions", and ``no errors". We note that these subjective evaluations focus on user experience and perceived control. Metrics like ``precise motions" and ``smooth motions" reflect how intuitive and seamless the system feels to users, rather than physical accuracy. This aligns with the primary goal of assistive systems: enhancing user experience and control. Given the multiple-trial, multiple-method nature of our study, we opted for this concise questionnaire instead of lengthier standardized assessments such as NASA-TLX to minimize participant burden and maintain engagement. The full questionnaire is available on our project website.

\begin{figure}[htbp]
\centering

\subfigure[Cereal pouring task]{
\begin{minipage}[t]{\columnwidth}
\centering
\includegraphics[width=\columnwidth]{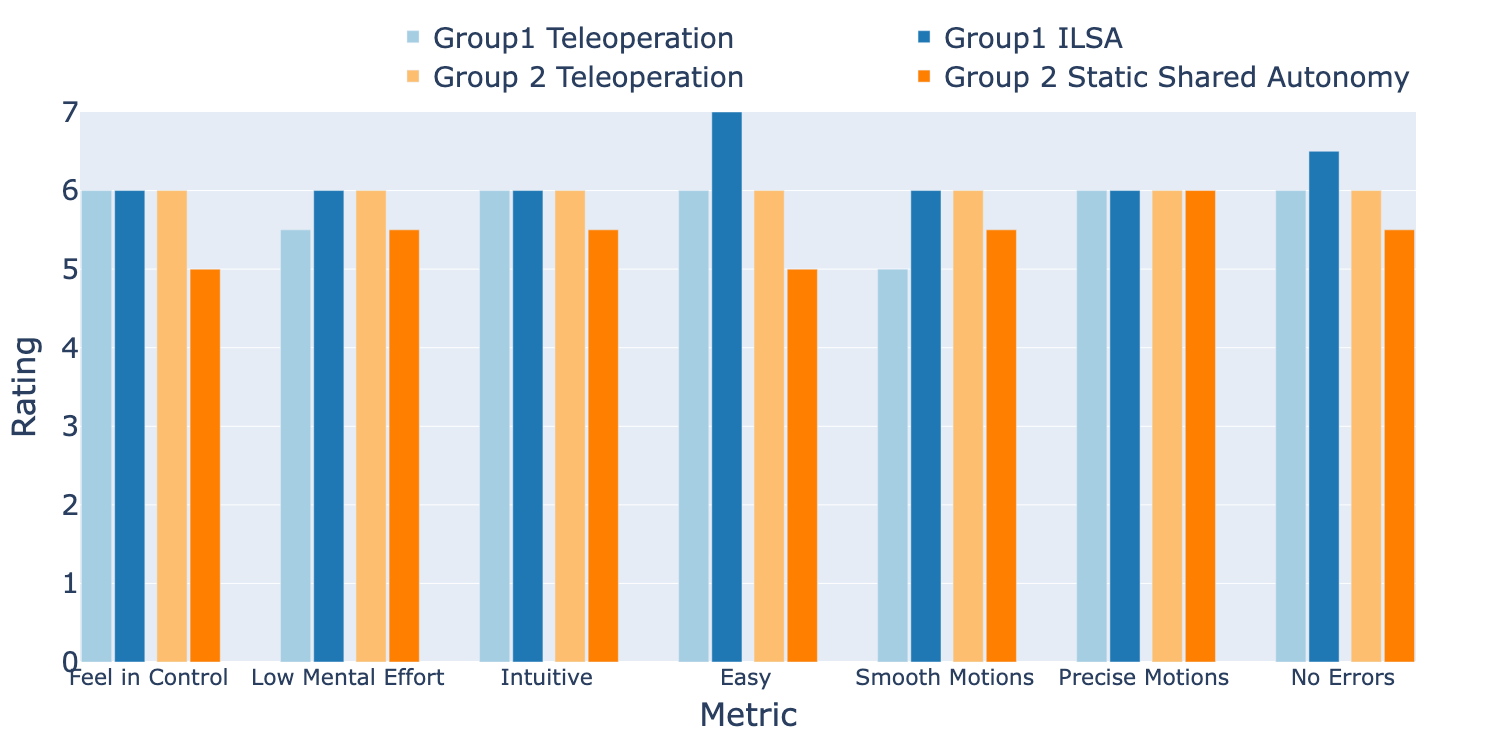}
\end{minipage}%
}%
\quad                
\subfigure[Pill bottle storage task]{
\begin{minipage}[t]{\columnwidth}
\centering
\includegraphics[width=\columnwidth]{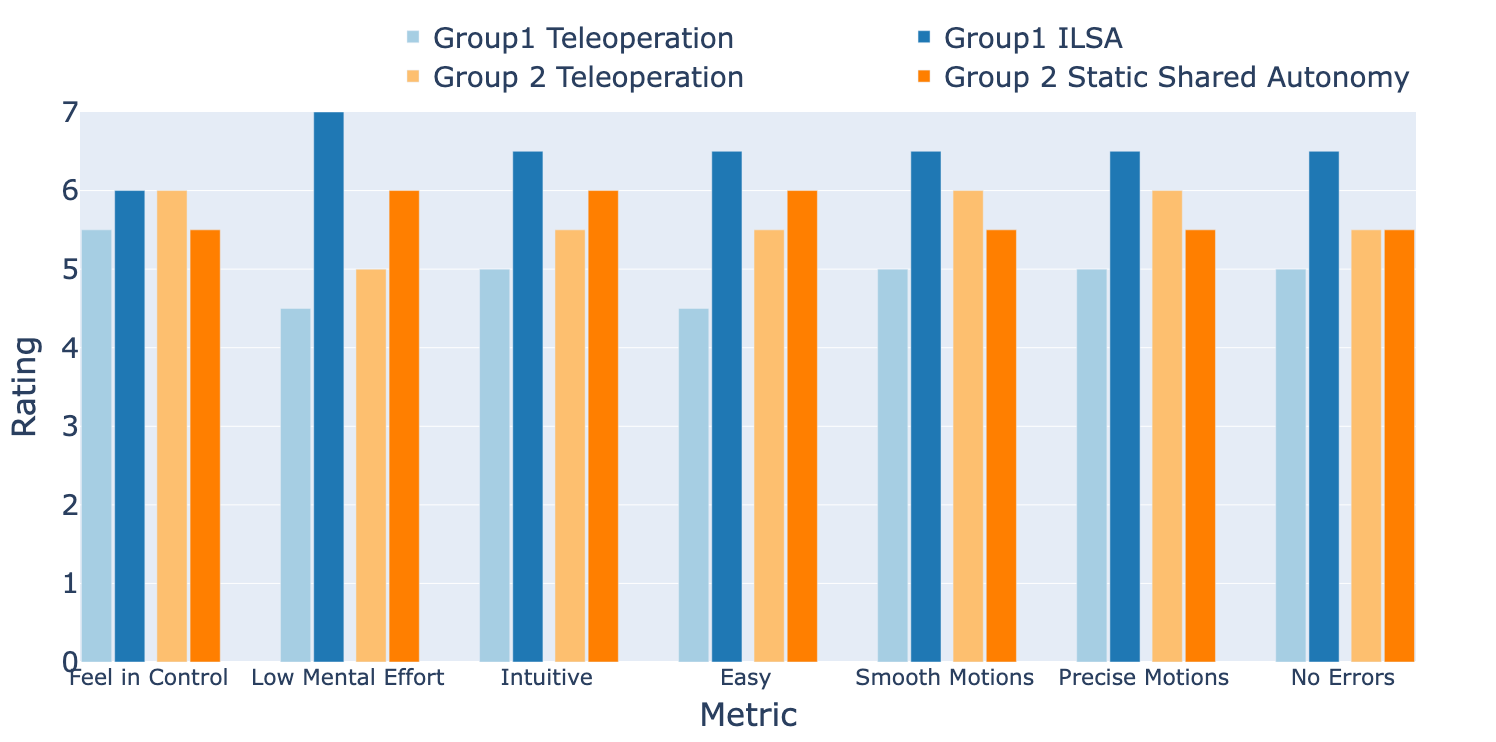}
\end{minipage}
}%

\centering
\caption{Median Likert item responses for all methods after trial 4.}
\label{fig:likert}
\vspace{-1em}
\end{figure}

Participants' median responses at trial 4 are shown in Fig.~\ref{fig:likert}. For both tasks, most participants reported a higher value on most metrics for ILSA compared to teleoperation. By applying a Wilcoxon signed-rank test, for the cereal pouring task, we observe a statistically significant difference between ILSA and teleoperation on ``low mental effort'' ($p<0.01$), ``intuitiveness'' ($p<0.05$), ``ease of task completion'' ($p<0.01$), ``smooth motions'' ($p<0.01$) and ``no errors'' ($p<0.05$). For the pill bottle storage task, we observe a statistically significant difference between ILSA and teleoperation on all 7 metrics ($p<0.05$ for ``feel in control'' and ``precise motions'', $p<0.01$ for ``low mental effort'', ``intuitiveness'', ``ease of task completion'', ``smooth motions'', and ``no errors''). These results support that users in general prefer ILSA over teleoperation, which provides evidence towards H1.

\section{CONCLUSIONS}
This paper presents ILSA, an Incrementally Learned Shared Autonomy system that continuously refines its assistive policy through user interactions. A key contribution of ILSA is its structured fine-tuning mechanism, which enables per-interaction adaptation by efficiently leveraging limited new data while retaining prior knowledge, ensuring a stable balance between adaptability and generalization. Through ablation and user studies, we demonstrate that ILSA effectively handles unforeseen real-world challenges, including those that fundamentally alter task dynamics, such as unexpected obstacles or spatial constraints, and outperforms alternative designs, teleoperation, and static shared autonomy.

\section*{NIST Disclaimer}
Certain commercial entities, equipment, or materials may be identified in this document in order to illustrate a point or concept. Such identification is not intended to imply recommendation or endorsement by NIST, nor is it intended to imply that the entities, materials, or equipment are necessarily the best available for the purpose.





\bibliographystyle{plain}
\bibliography{icra}

\appendices
\section{Implementation Details}
\label{appendix:implementation}

\subsection{Network Architecture}
\label{appendix:model_structure}
ILSA’s robot action generation model consists of a transformer encoder and three multilayer perceptrons (MLPs). Specifically, we adopt the transformer structure implemented in the Robomimic framework~\cite{robomimic}. The three MLPs each have two hidden layers with 1024 units per layer, followed by ReLU activations. All model components are trained jointly using the loss functions described below.

\subsection{Loss Functions}
\label{appendix:loss_functions}
Our loss function is defined as:
\begin{equation}
\mathcal{L} = \alpha \mathcal{L}_{demo, m} +  \beta \mathcal{L}_{demo, f} + \gamma \mathcal{L}_{direc} + \delta \mathcal{L}_{order}
\end{equation}
where $\alpha$, $\beta$, $\gamma$, and $\delta$ are weighting parameters empirically set to $\alpha = \beta = 1$ and $\gamma = \delta = 100$.

$\mathcal{L}_{demo, m}$ and $\mathcal{L}_{demo, f}$ are defined as follows:

\begin{equation}
\mathcal{L}_{demo, m} = \| a_{m,t} - a_{r,t} \|_2^2 
\end{equation}
\begin{equation}
\mathcal{L}_{demo, f} = \| a_{f,t} - a_{r,t} \|_2^2 
\end{equation}

where $a_{r,t}$ is the robot action in the training trajectories. These two loss terms supervise the intermediate and final outputs respectively, guiding the robot to take actions towards task completion. To further ensure that the final executed action aligns with the user's input, we introduce two additional loss terms, action direction loss $\mathcal{L}_{direc}$ and action ordering loss $\mathcal{L}_{order}$, to supervise the final output $a_{f,t}$. 

$\mathcal{L}_{direc}$ ensures that the signs of the robot action align with the user action in each dimension. It is defined as:
\begin{equation}
\label{direction}
\mathcal{L}_{direc} = 
 \sum_{i \in \{x, y, z\}} |a_{f, t, i}| \cdot 1_{\mathbb{R}_{<0}}(a_{f, t, i} \cdot a_{u, t, i})
\end{equation}
where $a_{f, t, i}$ and $a_{u, t, i}, i \in \{x, y, z\}$ denote the final output action and the user action in the $x$, $y$, $z$ dimensions, respectively. $1_{\mathbb{R}_{<0}}()$ is the indicator function for the set of negative real numbers.

The action ordering loss, $\mathcal{L}_{order}$, ensures that the relative magnitudes of the robot action align with the relative magnitudes of the user action across the $x$, $y$, and $z$ dimensions. It is defined as:

\begin{equation}
\label{order}
\begin{aligned}
\mathcal{L}_{order} = \sum_{(i,j) \in P} & \ ||a_{f, t, i}| - |a_{f, t, j}|| \cdot \\
& 1_{\mathbb{R}_{<0}}\Big( (|a_{f, t, i}| - |a_{f, t, j}|)(|a_{u, t, i}| - |a_{u, t, j}|) \Big)
\end{aligned}
\end{equation}
where $P=\{(x, y), (x, z), (y, z)\}$ represents the pairs of dimensions over which the relative magnitude differences are compared.

This term penalizes cases where the ordering of action magnitudes between the robot and user is inconsistent. For example, if the user's action along the $x$-dimension is greater than along the $y$-dimension, but the robot's corresponding action does not reflect this relationship, a penalty is applied. By minimizing this loss, the model is encouraged to maintain consistency with user's input.

\end{document}